\pdfoutput=1
\documentclass[11pt]{article}

\usepackage[final]{emnlp2021}
\usepackage{gb4e}
\noautomath
\usepackage{paralist}
\usepackage{mathtools}
\usepackage[inline]{enumitem}
\usepackage{times}
\usepackage{latexsym}
\usepackage{amsmath}
\usepackage[capitalize]{cleveref}
\usepackage{amssymb}
\usepackage{multirow,multicol}
\usepackage{graphicx}

\usepackage{array}
\usepackage{bm}
\usepackage{soul, color}
\usepackage{booktabs}
\usepackage[colorinlistoftodos]{todonotes}
\usepackage{enumitem}
\usepackage{amsthm}
\usepackage{caption}
\usepackage{subcaption}
\usepackage{tipa}
\usepackage{stackengine}
\usepackage{rotating}
\usepackage{tabularx}
\usepackage[T1]{fontenc}
\usepackage[utf8]{inputenc}

\usepackage{microtype}

\newcommand{\PreserveBackslash}[1]{\let\temp=\\#1\let\\=\temp}
\newcolumntype{C}[1]{>{\PreserveBackslash\centering}p{#1}}
\newcolumntype{R}[1]{>{\PreserveBackslash\raggedleft}p{#1}}
\newcolumntype{L}[1]{>{\PreserveBackslash\raggedright}p{#1}}

\DeclareMathOperator*{\ReLU}{\text{ReLU}}

\newcommand\code[1]{\texttt{#1}}

\crefformat{section}{\S#2#1#3} % see manual of cleveref, section 8.2.1
\crefformat{subsection}{\S#2#1#3}
\crefformat{subsubsection}{\S#2#1#3}
\title{STaCK: Sentence Ordering with Temporal Commonsense Knowledge}
\author{Deepanway Ghosal$^\dagger$, Navonil Majumder$^\dagger$,
  Rada Mihalcea$^\triangle$,
  Soujanya Poria$^\dagger$\\\\
  $^\dagger$ Singapore University of Technology and Design, Singapore\\
  $^\triangle$ University of Michigan, USA\\
  \texttt{deepanway\_ghosal@mymail.sutd.edu.sg}\\
  \texttt{\{navonil\_majumder, sporia\}@sutd.edu.sg}\\ \texttt{mihalcea@umich.edu} 
  }

\begin{document}
\maketitle
% \twocolumn[{%
% \renewcommand\twocolumn[1][]{#1}%
% \begin{center}
% \maketitle
%     \includegraphics[width=0.7\linewidth]{figs/example.pdf}
%     \captionof{figure}{\footnotesize Position of two sentences differ based on the dissimilar contextual utterances; the ordering is also inferred using commonsense knowledge in document A.}
%     \label{fig:example}
% \end{center}
% }]%

\begin{abstract}
Sentence order prediction is the task of finding the correct order of sentences in a randomly ordered document. Correctly ordering the sentences requires an understanding of coherence with respect to the chronological sequence of events described in the text. Document-level contextual understanding and commonsense knowledge centered around these events are often essential in uncovering this coherence and predicting the exact chronological order.  In this paper, we introduce STaCK---a framework based on graph neural networks and temporal commonsense knowledge to model global information and predict the relative order of sentences. Our graph network accumulates temporal evidence using knowledge of `past' and `future' and formulates sentence ordering as a constrained edge classification problem. We report results on five different datasets, and empirically show that the proposed method is naturally suitable for order prediction. % in academic paper abstracts, short stories, and visual captions.
The implementation of this work is publicly available at: \url{https://github.com/declare-lab/sentence-ordering}.
\end{abstract}

\section{Introduction}
%\hl{Soujanya: Tweak this section to showcase the novelty of our graph.}
Coherence is an essential quality for any natural language text \cite{Halliday76Cohesion}. Correct ordering of sentences is a necessary attribute of coherence. As such, there has been much research in correct sentence order detection due to its application in various down-stream tasks, such as, retrieval QA~\cite{wei2018fast}, multi-document summarization~\cite{10.5555/1622810.1622812}, automated content addition to text~\cite{mostafazadeh-etal-2016-corpus}, text generation~\cite{mostafazadeh-etal-2016-corpus}, and others. It also has potential applications in the evaluation of the quality of machine-generated documents. Existing approaches to sentence order prediction can be broadly classified into two categories: (1) sequence generation methods, and (2) pair-wise methods~\cite{zhu2021neural,prabhumoye2020topological}. While the former consider tagging the entire sequence, the latter takes one sentence pair at a time and predicts their relative order. Pair-wise methods ignore the importance of document level global information, i.e., while predicting the relative order of two sentences $(s_i, s_j)$, other sentences $s_k$ from the same document do not play any role. %The key limitation of the above Pair-wise methods is that global document-level information is not considered for the relative order prediction of sentence pairs. 
\begin{figure}[ht!]
    \centering
    \includegraphics[width=\linewidth]{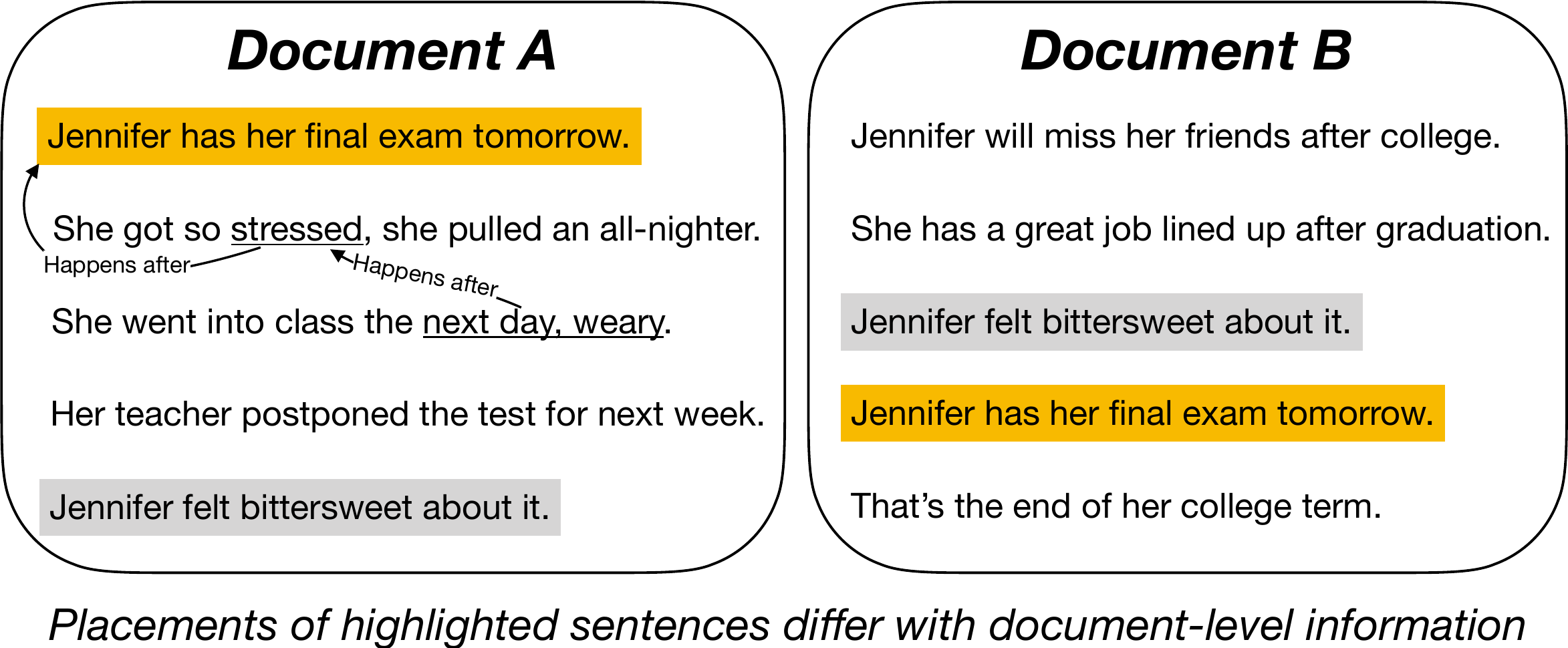}
    \caption{Position of two sentences differ based on the dissimilar contextual utterances; the ordering is also inferred using commonsense knowledge in document A.}
    \label{fig:example}
  %  \vspace{-0.3cm}
\end{figure}
Global document information is especially important while predicting the relative order of sentences that are further apart, as it can provide essential contextual cues. As an example, consider the two highlighted sentences in the two sample documents shown in Figure \ref{fig:example}. Although the sentences describe seemingly identical events, they have a different relative order in the two documents because of their different contexts. We recognize this fundamental limitation in existing methods, and we hypothesize that global information is essential for predicting the relative order of a sentence pair. It encompasses not only the semantic information of the discourse, but also commonsense knowledge centered around all the sentences of the document. %We surmise that such information is often necessary in predicting the correct order of the sentences. This is empirically shown through various analytical studies in \cref{sec:results}. 

In this paper, we propose a graph-based framework to represent sentences in a document and their relations. Using a a two-layer relational graph convolutional network (RGCN) applied on this graph, we build a classifier that is able to learn the relative order of sentences in a document by accounting for the global document information encoded in the graph. 
%for information propagation across the constituent nodes. The relative order of a sentence pair is determined using a classifier that selects exactly one of the two possible orders.% In our work, we firstly construct a complete directed graph with nodes representing individual sentences in a paragraph.  Furthermore, a global node representing the entirety of the unordered document is connected to all of the sentence nodes to facilitate some contextual information flow across sentences. This graph is fed to a two-layer relational graph convolutional network (RGCN) for information propagation across the constituent nodes. The relative order of a sentence pair is determined using a classifier that selects exactly one of the two possible orders. The edge representing this correct order is kept for all possible sentence pairs and the sentence nodes are topologically sorted concerning these edges to obtain the final ordering of sentences. Our constructed graph ensures the use of document-level information in the task. It should be noted sequence generation methods such as \cite{kumar2020deep} also use document-level information. From the empirical results, it is evident that our method is superior in capturing the enriched document information by the means of a graph neural network.
We further infuse commonsense knowledge (CSK) information into this graph to improve the model performance. The key motivation is that temporal commonsense knowledge can bring important information about events that may occur before or after an event described in a sentence. %We assume that the overlap between such temporal commonsense knowledge and the content of other sentences may induce an overall ordering of sentences. Two nodes representing possible past and future events of a sentence, as per CSK, are thus connected to each respective sentence node in the graph explained in the last paragraph. We empirically show that this approach outperforms the state-of-the-art method on various datasets.

Our paper makes two important contributions. Firstly, we show how we can construct a document graph that captures global context information about the document. We employ a RGCN to encode the information in this graph and build an edge classifier that  predicts the relative order of sentence pairs. Unlike previous work attempting to predict the relative order of sentence pairs, our approach explicitly accounts for global document-level information. Secondly, we infuse \emph{temporal} commonsense knowledge into our graph convolutional neural network to further improve its performance. To the best of our knowledge, there is no prior work that attempted the use of CSK for sentence order prediction. Our results suggest that the graph representation encoding global document information and the temporal CSK are both effective to determine the order of sentences.

\textbf{Sequence generation vs pair-wise methods for sentence ordering:}
In the literature, the sentence ordering task is also addressed using generative modeling~\cite{chowdhury2021reformulating} where the goal is to generate the correct positions of the sentences of
% randomly shuffled sentence sequence 
a shuffled document as an integer sequence. In contrast, our method first classifies the sentence pairs followed by topological ordering. Despite achieving impressive performance, there exist some key differences that make the generative approaches \texttt{fundamentally different and incomparable} from the family of sentence pair classification-based or pair-wise approaches:
\begin{enumerate}[itemsep=0ex,leftmargin=*]
    \item The generative models for sentence ordering take sequences as input that contains temporal information in the form of learned positional embeddings. One could argue that this temporal information is noisy and thus would not provide any useful information to the model. However, this does not remove all the temporal information from the input that could assist the model, e.g., a shuffled order ${5,1,3,4,2}$ still contains valid temporal sequence or information i.e., sentence $1$ precedes sentence $4$, and $2$, sentence $3$ precedes sentence $4$. Hence, a sequence generation model that accepts positional encoding of the sentences can still get confounding temporal signal despite the shuffling. %It is worth investigating whether such a problem formulation is similar in nature to denoising autoencoders.
    \item The generative models for sentence ordering tend not to work when the sentence count during inference exceeds the highest sentence count observed in training. 
    % We have faced this issue in our experiments with the generative models e.g., if we train the model with a maximum of five sentences, during inference the model does not generate $6$ in the output. 
    For instance, we found in our experiments that generative models trained on samples with five sentences would only generate tokens $1-5$ during inference, even if the test document has more sentences. This raises serious questions about such models' reasoning ability in zero-shot situations. One future direction to tackle such issues would be passing the input sequence length as a prompt or input to the generative model. The use of sequence length embedding could also be a possible solution. Contrary to this, pair-wise methods are robust at handling any number of sentences.
\end{enumerate}

\section{Background}
\label{sec:related_work}
Coherence and the problem of sentence order prediction have been extensively studied in literature due to their applicability in various downstream problems. Early work in this direction mainly used domain knowledge and handcrafted linguistic features to model the relation between sentences in a document~\cite{lapata2003probabilistic,barzilay2004catching,barzilay2008modeling}. Sentence ordering methods in recent literature are primarily based on neural network architectures, and can be broadly categorized into two main families - i) Sequence generation methods and ii) Pair-wise methods.  

Sequence generation methods use the entire sequence of the randomly ordered document to model local and global information. This information is then used to predict the correct order. The sentences and documents are typically encoded using a recurrent or transformer-based network ~\cite{gong2016end,yin2020enhancing,kumar2020deep}. Hierarchical encoding schemes are also common ~\cite{wang2019hierarchical}. Prediction is then generally performed with a pointer network decoder ~\cite{vinyals2015pointer} based on beam search. Alternatively, ranking losses~\cite{xia2008listwise} have also been explored to circumvent the expensive beam search algorithm ~\cite{kumar2020deep}. Such models predict a score for each sentence, which are then sorted to obtain the final order.
The Pair-wise methods are motivated by a different principle of sentence ordering. These models aim to predict the relative order of each pair of sentences in the document. The final order is then constrained on all of the predicted relative orders. 
% Inferring the final order requires solving the constraint conditions imparted by the relative orders. 
The constraint solving problem is generally tackled with topological sorting ~\cite{prabhumoye2020topological}, or more sophisticated neural network models ~\cite{zhu2021neural}. 

Our proposed STaCK framework falls under this family of Pair-wise models.
% The proposed model in this work comes under the umbrella of Pair-wise methods. We surmise that temporal commonsense knowledge and global document information are elemental for predicting the relative order of a sentence pair. This is one of the key limitations of the state-of-the-art Pair-wise methods ~\cite{prabhumoye2020topological} and ~\cite{zhu2021neural}. 
In STaCK, temporal commonsense is modelled using the Commonsense Transformers (COMET) model~\cite{hwang2020comet}. The COMET model uses a BART~\cite{lewis2020bart} sequence-to-sequence encoder decoder framework and is pretrained on the $\text{ATOMIC}_{20}^{20}$ commonsense knowledge graph~\cite{hwang2020comet}. The pretraining objective is to take a triplet $\{s, r, o\}$ from the knowledge graph, and generate the object phrase $o$ from concatenated subject and relation phrase $s$ and $r$. The set of relations $\mathcal{R}$ include temporal relations `is before' and `is after'. COMET is pretrained on approximately 50,000 of such temporal triplets along with other commonsense relations from $\text{ATOMIC}_{20}^{20}$. The pretraining on commonsense knowledge ensures that COMET is capable of distinguishing cause-effects (causality), past-future (temporality), and other event-centered commonsense knowledge.
\section{Methodology}
\begin{figure*}[t]
    \centering
    \includegraphics[width=0.9\linewidth]{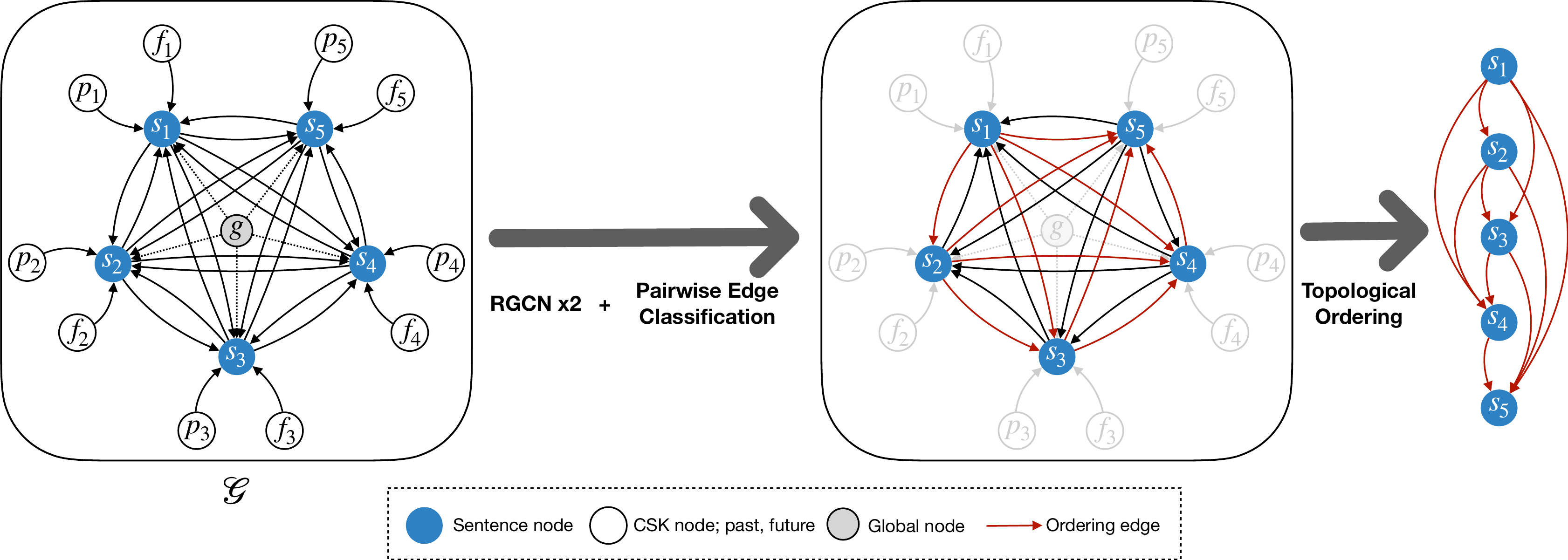}
    \caption{\footnotesize Illustration of STaCK.}
    % \vspace{-0.35cm}
    \label{fig:model}
\end{figure*}

An overall illustration of the proposed STaCK framework is shown in \cref{fig:model}.
% Let the document/story $\mathcal{D}$ have $n$ sentences in it. 
We represent document $\mathcal{D}$ consisting of $n$ sentences as a directed graph $\mathcal{G = (V, E, R)}$, with nodes $v_{i} \in \mathcal{V}$ and directed labeled edges $(v_i, r_{ij}, v_j)\in \mathcal{E}$, where $r_{ij} \in \mathcal{R}$ is the relation type of the edge between $v_i$ and $v_j$. Initial node embeddings are denoted as $g_i$.
\subsection{Graph Construction} 
\label{sec:graph}
The graph is constructed from the given document $\mathcal{D}$ as follows:
\paragraph{Nodes and Node Embeddings:} We consider three different types of nodes in $\mathcal{V}$:
%\begin{itemize}
%\setlength\itemsep{0em}

\noindent \textbf{Sentence nodes}. Each sentence $s_i$ in $\mathcal{D}$ is a sentence node in the graph.
We pass the sentence through a DeBERTa ~\cite{he2020deberta} model and use the final layer vector corresponding to starting token \code{<s>} as the node embedding.

\noindent\textbf{CSK nodes}. For each sentence $s_i$, we have a past and future node $p_i$ and $f_i$, respectively.
The CSK node embeddings are initialized from the BART encoder of the COMET model. 
% COMET uses a BART encoder to encode sentences with respect to commonsense relations. 
Following 
% the methodology described in 
COMET, we append temporal relation specific tokens \code{isAfter [GEN]} and \code{isBefore [GEN]} with the sentence $s_i$. The concatenated text sequence is passed through the BART encoder and final layer vector corresponding to \code{<s>} is used as the node embedding for $p_i$ and $f_i$. 
% We also tried adding other temporal and causal commonsense relations such as \emph{causes}, \emph{As-A-result}, \emph{desires}, \emph{requires} as nodes to the graph. However, they did not result in any significant performance improvements. We posit this could be due to the fact that there exists a large overlap between the generated output of \emph{isBefore}, \emph{isAfter} and the four relations as mentioned above. Nonetheless, we think all types of CSK relations available in COMET can be used in the task and the required graph structure to accommodate those additional CSK is left for the future work.

\noindent\textbf{Global node}. The entire document $\mathcal{D}$ is considered as an additional global node $g$ in $\mathcal{G}$.
We pass the document through a non-positional (position embeddings removed) RoBERTa model~\cite{liu2019roberta}, and use the final layer vector corresponding to starting token \code{<s>} as the global node embedding. This non-positional model is insensitive to the sequence of tokens passed into it. 
% In other words, the embedding of the global node depends only upon the tokens present in $\mathcal{D}$, and not their order. 
The usage of a non-positional model is essential, as the model must not have any information about the relative order of the sentences. For a document $\mathcal{D}$ consisting $n$ sentences, we have $3n+1$ total nodes in $\mathcal{G}$.

\paragraph{Edges and Relations:}
We construct edges with different relations based on the constituent nodes:
%\begin{itemize}
% \setlength\itemsep{0em}
    \textbf{ Sentence edges}. Each sentence node $s_i$ in $\mathcal{D}$ is connected to all other sentence nodes $s_j$ (where $j \neq i$) in $\mathcal{D}$ with relation $r_s$. The directed edge is denoted as $(s_i, r_s, s_j)$. Our formulation leads to bidirectional edges between each sentence pair, i.e. both $(s_i, r_s, s_j)$ and $(s_j, r_s, s_i) \in \mathcal{E}$.
    
    \noindent\textbf{CSK edges}. Each CSK node $p_i$ and $f_i$, has an edge with the corresponding sentence node: $(p_i, r_p, s_i)$, and $(f_i, r_f, s_i)$. The relation is set different for past and future CSK nodes. The direction of the edge is from the CSK node to the sentence node.
    
    \noindent \textbf{Global edges}. The global node $g$ has an edge with every sentence node: $(g, r_g, s_i)$, for $i=1,2,..,n$. As indicated, the direction of the edge is taken from the global node to the sentence node.
    % \item Self edges: Each node $v$ in the graph has an edge with itself to ensure self-dependent and direct information propagation in the graph neural network framework. These edges are denoted as $(v, r_{self}, v)$, where $v = \{s_i, p_i, f_i, g\}$ for $i=1,2,..,n$.
%\end{itemize}

\subsection{Graph Encoder: RGCN}
\label{sec:gcn}
We use a two-layer Relational Graph Convolutional Network (RGCN)~\cite{schlichtkrull2018modeling} as our graph encoding model. The RGCN model is able to accumulate relational evidence
in multiple inference steps from the neighborhood around a given node. The RGCN model is a natural choice of encoding algorithm as it enables the modelling of different relations across our graph. In RGCN, the transformation of node embeddings are performed as follows: $f(\mathbf{x}_i , l) = \ReLU (\sum\limits_{r \in \mathcal{R}}^{} \sum\limits_{j \in N_{i}^{r}}^{} \frac{W_{r}^{(l)} \mathbf{x}_{j}}{c_{i,r}} + W_{0}^{(l)}\mathbf{x}_{i}), \mathbf{h}_{i} = \mathbf{h}_{i}^{(2)} = f(\mathbf{h}_i^{(1)},2); \mathbf{h}_{i}^{(1)} = f(\mathbf{g}_i \, 1)$.
%

% {\scriptsize
% \begin{align*}
%     f(\mathbf{x}_i , l) &= \ReLU (\sum\limits_{r \in \mathcal{R}}^{} \sum\limits_{j \in N_{i}^{r}}^{} \frac{W_{r}^{(l)} \mathbf{x}_{j}}{c_{i,r}} + W_{0}^{(l)}\mathbf{x}_{i}), \\
%     \mathbf{h}_{i} &= \mathbf{h}_{i}^{(2)} = f(\mathbf{h}_i^{(1)},2); \quad \mathbf{h}_{i}^{(1)} = f(\mathbf{g}_i \, 1),
% \end{align*}
% }%
\noindent where $N_{i}^{r}$ indicates the neighbouring nodes of $v_i$ under relation $r \in \mathcal{R}$; $c_{i,r}$ is a normalization constant which either can be learned in a gradient-based learning setup, or can be set in advance, such that, $c_{i,r} = |N_{i}^{r}|$ and $W_{r}^{(1/2)}$, $W_{0}^{(1/2)}$ are learnable parameters of the transformation. The self-dependent connection with weight $W_{0}^{(1/2)}$ is added to ensure direct information among same nodes in consecutive layers in the graph framework.

For node $v_i$, we start with initial node embeddings $g_i$ (\cref{sec:graph}), and transform it to $h_i$, following the two layer RGCN transformation process.

\subsection{Graph Decoder: Pairwise Edge Classifier}
The final module in our graph network is built upon the principle of pairwise edge classification. This module predicts the relative order between any two sentences in $\mathcal{D}$ by using the initial input embeddings $g$ and output activations $h$ from the RGCN encoder. 
For example, let us take two sentences $s_i$ and $s_j$, where $i<j$, i.e. $s_i$ appears earlier in $\mathcal{D}$, and $s_j$ appears later. In this formulation, we will first consider the bidirectional edges between $s_i$ and $s_j$ in $\mathcal{E}$ --- $(s_i, r_s, s_j)$ and $(s_j, r_s, s_i)$. The classification objective is then to classify the first edge $(s_i, r_s, s_j)$ as 1 and second edge $(s_j, r_s, s_i)$ as 0. In other words, if the originating sentence of the directed edge appears earlier than the destination sentence in the original document, then we predict the class of the edge as 1, otherwise 0.

To achieve this, we use a function $f$, which takes the concatenated feature vectors $m_i = [g_i, h_i]$ and $m_j = [g_j, h_j]$ and outputs a single scalar value as a score. We compute $f(m_i, m_j)$, and $f(m_j, m_i)$ and normalize them with softmax activation to output two probabilities $p_{ij}, p_{ji}$ for the two edges $(s_i, r_s, s_j)$ and $(s_j, r_s, s_i)$. The softmax operation ensures that, $p_{ij}+p_{ji}=1$. During training, the probabilities are pushed towards 1 and 0 for the paired edges. During inference, for sentences $s_x$ and $s_y$, if $p_{xy}$ > $p_{yx}$ then we predict $s_x$ appears earlier than $s_y$ ($s_x \rightarrow s_y$), or vice versa ($s_x \leftarrow s_y$).

Naturally the function $f$ has to be sensitive to the order of its inputs in this formulation. The more different the outputs scores are, the more the normalized probabilities are pushed towards 0 and 1. From our experiments, we find that functions that have an anti-symmetric component are most suitable for $f$. In particular we use, $f(m_i, m_j) = w^T \sin(m_i - m_j)$, where $m_i, m_j \in \mathbb{R}^d$ and the sine operation is performed element-wise. $w \in \mathbb{R}^d$ is the learnable parameter of the function. The sine operation is the anti-symmetric component in our function, as, 
$\sin(m_i - m_j) = - \sin(m_j - m_i)$. Other functions such as outer product performed worse than the sine function in our experiments.

\subsection{Topological Sorting}
The topological sorting method ~\cite{prabhumoye2020topological} is used to obtain the final ordered sequence of sentences from the all the pairwise classifications. If the pairwise classifier predicts that $p_{xy}$ > $p_{yx}$ i.e. $s_x \rightarrow s_y$, then the sorting method ensures $s_x$ comes before $s_y$ in the final ordering $\hat{o}$. 
% If there are cycles in the pairwise predictions, then a random spanning tree is generated, on which the topological sort can be applied.
\begin{table}[t]
\small
\centering
\resizebox{0.8\linewidth}{!}{
\begin{tabular}{lrrrrrr}
\toprule
Dataset & Min & Max & Avg & Train & Val & Test \\
\midrule
NeurIPS & 1 & 15 & 6 & 2448 & 409 & 402 \\
AAN & 1 & 20 & 5 & 8569 & 962 & 2626 \\
NSF & 2 & 40 & 8.9 & 96017 & 10185 & 21573 \\
SIND & 5 & 5 & 5 & 40155 & 4990 & 5055 \\
ROCStory & 5 & 5 & 5 & 78529 & 9816 & 9816 \\
\bottomrule
\end{tabular}
}
\caption{\footnotesize Statistics of the datasets. Min, Max, and Avg indicates minimum, maximum, and average number of sentences in a document. Train, Val, Test indicates number of documents in those splits respectively.}
\label{tab:dataset}
\end{table}
\section{Experimental Setup}
\begin{table*}[t]
\small
\centering
\resizebox{0.86\linewidth}{!}{
\begin{tabular}{l|cr|cr|cr|cr|cr}
\toprule
\multirow{2}{*}{\textbf{Method}} & \multicolumn{2}{c|}{\textbf{NeurIPS}} & \multicolumn{2}{c|}{\textbf{AAN}} & \multicolumn{2}{c|}{\textbf{NSF}} & \multicolumn{2}{c|}{\textbf{SIND}} & \multicolumn{2}{c}{\textbf{ROCStory}} \\
%\cline{2-11}
& $\tau$ & PMR & $\tau$ & PMR & $\tau$ & PMR & $\tau$ & PMR & $\tau$ & PMR \\
\midrule
LSTM Pointer Net & 0.7373 & 20.95 & 0.7394 & 38.30 & 0.5460 & 10.68 & 0.4833 & 12.96 & 0.6787 & 28.24\\
Hierarchical Attention Net & 0.7008 & 19.63 & 0.6956 & 30.29 & 0.5073 & 8.12 & 0.4814 & 11.01 & 0.6873 & 31.73 \\
Sentence Entity Graph & 0.7370 & 24.63 & 0.7616 & 41.63 & 0.5602 & 10.94 & 0.4804 & 12.58 & 0.6852 & 31.36\\
ATTOrderNet TwoLoss & 0.7357 & 23.63 & 0.7531 & 41.59 & 0.4918 & 9.39 & 0.4952 & 14.09 & 0.7302 & 40.24 \\
RankTxNet ListMLE & 0.7462 & 24.13 & 0.7748 & 39.18 & 0.5798 & 9.78 & 0.5652 & 15.48 & 0.7602 & 38.02 \\
B-TSort & 0.7824 & 30.59 & 0.8064 & 48.08 & 0.4813 & 7.88 & 0.5632 & 17.35 & 0.7941 & 48.06 \\
% B-TSort DeBerta & 0.8003 & 32.09 & 0.8193 & 49.23 & & & 0.6114 & 21.54 & & \\
Constraint Graphs & 0.8029 & 32.84 & 0.8236 & 49.81 & 0.6082 & \textbf{13.67} & 0.5856 & 19.07 & 0.8122 & 49.52 \\
\midrule 
STaCK w/o CSK Nodes, Edges & 0.8035 & 33.67 & 0.8365 & 51.10 & 0.6567 & 11.79 & 0.6154 & 19.84 & 0.8391 & 53.04 \\
\midrule
STaCK & \textbf{0.8166} & \textbf{37.31} & \textbf{0.8556} & \textbf{54.01} & 0.6582 & 12.26 & \textbf{0.6194} & \textbf{20.79} & \textbf{0.8534} & \textbf{55.96} \\
\quad same CSK Edges: $r_p = r_f$ & 0.8146 & 37.06 & 0.8472 & 52.25 & \textbf{0.6642} & \textbf{12.41} & 0.6172 & 20.03 & 0.8470 & 54.32 \\
% \quad w/o CSK Nodes, Edges & 0.8035 & 33.30 & 0.8365 & 51.10 & 0.6567 & 11.79 & 0.6154 & 19.84 & 0.8391 & 53.04 \\
\bottomrule
\end{tabular}
}
\caption{\footnotesize Comparison of results of our model against various methods. Scores are reported at best validation $\tau$. The performance difference between STaCK (w/ and w/o csk) and the baselines are statistically significant according to paired t-test with p < 0.05. }
\label{tab:results}
\end{table*}
\subsection{Datasets}
We benchmark the proposed STaCK framework on five different datasets. 
% following recent literature. 
\textbf{NeurIPS, AAN, NSF Abstracts.}
These three datasets contain abstracts from NeurIPS papers, AAN papers, and the NSF Research Award abstracts respectively ~\cite{logeswaran2018sentence} . The number of sentences in each abstract varies significantly, ranging from two to forty.
\textbf{SIND.} This is a sequential vision to language dataset~\cite{huang2016visual}  used for the task of visual storytelling. Each story contains five images and their descriptions in natural language.
\textbf{ROCStory.} ~\cite{mostafazadeh-etal-2016-corpus} introduce a dataset of short stories capturing a rich set of causal and temporal commonsense relations between daily events. The dataset has been used for evaluating story understanding, generation, and script learning. All stories have five sentences. For ROCStory, we use the train, val, test split as used in ~\citet{zhu2021neural}. For the other datasets, we use the splits following the original papers. Some statistics about the datasets is shown in \cref{tab:dataset}.

\subsection{Evaluation Metrics}
% Following previous work, we adopt two commonly used metrics -- Kendall's $\tau$ and Perfect Match Ratio (PMR) to benchmark our proposed model.
\textbf{Kendall's \boldsymbol{$\tau$}} is an automatic metric widely used for evaluating text coherence. It measures the distance between the predicted order and the correct order of sentences in terms of number of inversions. It is calculated as, $\tau = 1 - 2I/ \binom n2$, where I is the number of pairs predicted with incorrect relative order and n is the number of sentences in the paragraph. The score ranges from -1 to 1, with 1 indicating perfect prediction order.
\textbf{PMR} (Perfect Match Ratio) measures the percentage of instances for which the entire order of the sequence is correctly predicted. It is a more strict metric, as it only gives credit to sequences that are fully correct, and does not give  credit to partially correct sequences.

\subsection{Training Setup}
Training is performed by optimizing the binary cross-entropy loss function for pairwise edge classification. We use the AdamW optimizer with a learning rate of 1e-6 for the parameters of the transformer models used in extracting node embeddings. For the parameters of the RGCN encoder and edge classifier, we use the Adam optimizer with a learning rate of 1e-4. We train our models for 10 epochs with a batch size of 8 documents. Test results are reported corresponding to the best validation $\tau$.

\section{Results and Analysis}
\label{sec:results}
\begin{table}[t]
%\small
\centering
\resizebox{\linewidth}{!}{
%{
\begin{tabular}{l|ccccc|ccccc}
\toprule
\multirow{2}{*}{\textbf{Method}} & \multicolumn{5}{c|}{\textbf{STaCK}} & \multicolumn{5}{c}{\textbf{STaCK w/o CSK}} \\
& First & Last & Abs & LCS & D-Win=1 & First & Last & Abs & LCS & D-Win=1\\
\midrule
NeuRIPS & \textbf{93.3} & \textbf{78.8} & \textbf{63.6} & \textbf{83.1} & \textbf{87.5} & 92.2 & 76.1 & 59.9 & 81.1 & 86.3\\
AAN & \textbf{93.2} & \textbf{82.6} & \textbf{71.6} & \textbf{86.6 }& \textbf{92.0} & 91.9 & 79.9 & 68.0 & 85.1 & 90.5 \\
NSF & \textbf{86.2} & 56.9 & \textbf{33.9} & \textbf{65.6} & \textbf{58.7} & 85.4 & \textbf{58.0} & 33.6 & 65.4 & 58.3 \\
SIND & \textbf{83.2} & \textbf{66.3} & \textbf{54.2} & \textbf{77.4} & \textbf{84.1} & 82.7 & 65.6 & 53.1 & 76.9 & 83.5 \\
ROCStory & \textbf{96.1} & \textbf{82.1} & \textbf{76.7} & \textbf{89.2} & \textbf{95.2} & 95.6 & 81.7 & 76.1 & 88.9 & 94.8 \\
\bottomrule
\end{tabular}
}
\caption{\footnotesize Performance of models with and without commonsense knowledge. We report accuracy of predicting first, last, and absolute (Abs) position of sentences correctly. Longest common subsequence (LCS) ratio and displacement within window 1 (D-Win=1) metric are also reported in percentage.}
\label{tab:csk_compare}
\end{table}
\subsection{Baselines and State-of-the-art Methods}
We compare STaCK against the following methods:
\noindent \textbf{LSTM Pointer Net}~\cite{gong2016end}: A word2vec embedding based LSTM model with pointer network decoder.
\noindent \textbf{Hierarchical Attention Net}~\cite{wang2019hierarchical}: A word and sentence based hierarchical network with LSTMs and multihead attention encoder coupled with a multihead attention decoder.
\noindent \textbf{Sentence Entity Graph}~\cite{yin2019graph}: A model based on sentence entity graph with graph recurrent network encoder and pointer network decoder.
\noindent \textbf{ATTOrderNet TwoLoss}~\cite{yin2020enhancing}: An improved version of Attention Order Net with a pointer network decoder enhanced by two pairwise ordering prediction modules
\noindent \textbf{RankTxNet ListMLE}~\cite{kumar2020deep} uses a BERT sentence encoder and predicts a score for each sentence which are sorted to obtain to sentence order. The ListMLE function~\cite{xia2008listwise} is used as the objective function.
\noindent \textbf{BERT Topological Sort (BT-Sort)}~\cite{prabhumoye2020topological}: A pairwise model which applies the BERT-base encoder on concatenated sentence pairs to predict the relative order. For sentence pair (s1, s2), the input to the BERT encoder is \emph{<CLS> s1 <SEP> s2 <SEP>}, and the classification is performed from the \emph{<CLS>} token vector. Topological sort is then used to obtain the final prediction order from all relative orders. 
\noindent \textbf{Constraint Graphs}~\cite{zhu2021neural}: Another pairwise model which also applies the BERT-base encoder on concatenated sentence pairs to predict the relative order of the sentences. Constraints from the relative orders are then represented as a constraint graphs and
integrated into the sentence representation by using Graph Isomorphism Networks (GINs). All sentence representations from the GINs are fused together to predict the final score of sentences. The ListMLE objective function~\cite{xia2008listwise} is then used on these scores to predict the final order.

Among the above methods, \textbf{BT-Sort} and \textbf{Constraint Graphs} are of our main interest for comparative study. Constraint Graphs model is the current state-of-the-art for sentence order prediction.

\subsection{Main Results}
\cref{tab:results} shows the results across all the datasets for the baseline methods and our proposed model. STaCK achieves improved scores over the previous state-of-the-art across almost all the datasets on both evaluation metrics. Interestingly, we observe that the improvement in the $\tau$ metric is more significant in NSF and SIND. However, for NeuRIPS, AAN, and ROCStory the improvement in more prominent in the PMR metric. A modification of model which don't use any commonsense knowledge (\emph{STaCK: w/o CSK Nodes, Edges}) also surpass previous state-of-the-art results in most cases. We expand upon the obtained results and report a number of analysis studies next.

\subsection{Novelty of the Proposed Graph-based Model with CSK Nodes and Edges}
% \hl{highlight this section. Write clearly how can we fairly compare to the sota.}
State-of-the-art models BT-Sort and Constraint Graphs use sentence pair concatenation method to perform the relative order prediction between a pair of sentences. This method is widely used in GLUE style classification tasks. As illustrated before, this method doesn't consider any document level information for the relative order prediction. We also compare our proposed graph model without any CSK nodes and edges to the state-of-the-art methods. We report the results for this model in Table 2 in row \emph{STaCK: w/o CSK Nodes, Edges}. It can be observed that even after removing the CSK components, our graph model achieves improved scores in all datasets except the PMR metric in NSF. BT-Sort and our proposed model uses the same topological sort method to infer the final order of sentences. The significant improvement of STaCK: w/o CSK Nodes, Edges over BT-Sort can thus be directly attributed to the integration of document level information in our graph. Furthermore, even though Constraint Graphs uses a parametric neural network model to infer the final order of the sentences (compared to non-parametric topological sort of ours), it records an overall poorer performance across most metrics. From the empirical results, we conclude that document level information is indeed crucial for the task of sentence order prediction. In the future, the topological sorting employed in our work can be replaced with a more complex neural network-based sorting approach as used in the Constraint Graphs by \cite{zhu2021neural}.
% compared to \emph{Ours: Graph w/o CSK}.
A natural question might arise --- what if we use a different transformer encoder for the state-of-the-art models? We find that this change doesn't improve the results of the state-of-the-art models due to a mismatch in pretraining objective functions and the GLUE style classification setting. Other choices of encoders such as RoBERTa, ALBERT, or DeBERTa perform poorly compared to BERT for both BT-Sort and Constraint Graphs~\cite{zhu2021neural}. These encoders are not pretrained with the next sentence prediction (NSP) objective used in BERT. The NSP objective is similar to the concatenated sentence pair classification strategy used in BT-Sort and Constraint Graphs, enabling BERT to obtain the best possible performance.

% Similar conclusions have also been reported in Constraint Graphs~\cite{zhu2021neural}. 

% We expand upon this analysis in \cref{sec:encoder_choice}.
% BT-Sort and Constraint Graphs are the current state-of-the-art models which use a BERT encoder. Both the models come under the family of pair-wise models (\cref{sec:related_work}), which first predict the relative order between each pair of sentences in the document. The final order is then inferred from all the relative orders. The relative order prediction is performed by concatenating the sentence pairs with a \emph{<SEP>} token and then passing through BERT encoder. This setting directly aligns with the NSP objective of BERT, and is capable of achieving state-of-the-art results. However, as reported in ~\cite{zhu2021neural}, replacing the BERT encoder with a RoBERTa encoder results in poorer performance because of the absence of the NSP objective. Interestingly, using an ALBERT encoder also results in a performance drop, even though ALBERT was pre-trained with a sentence order prediction objective (albeit rather differently). Furthermore, we also experimented by replacing the the BERT encoder of BT-Sort with DeBERTa and found that the performance does not surpass the reported results of BERT. 
\begin{table}[t]
 \small
\centering
\resizebox{0.8\linewidth}{!}{
\begin{tabular}{l|cc|cc|cc}
\toprule
\multirow{2}{*}{\textbf{Method}} & \multicolumn{2}{c|}{\textbf{BT-Sort}} & \multicolumn{2}{c|}{\textbf{CG}} & \multicolumn{2}{c}{\textbf{Ours}} \\
& F & L & F & L & F & L \\
\midrule
AAN & 89.5 & 79.8 & 91.5 & 80.4 & \textbf{93.2} & \textbf{82.6}\\
SIND & 78.1 & 58.4 & 79.8 & 60.4 & \textbf{83.2} & \textbf{66.3} \\
NeuRIPS & 89.8 & 75.1 & - & - & \textbf{93.3} & \textbf{78.8} \\
NSF & - & - & - & - & 86.2 & 56.9 \\
ROCStory & - & - & - & - & 96.1 & 82.1\\
\bottomrule
\end{tabular}
}
\caption{\footnotesize Test accuracy of predicting the first (F) and last (L) sentences correctly. \textbf{CG}: Constraint Graphs model. Results are reported for BT-Sort and Constraint Graphs models wherever available.}
\label{tab:fl}
\end{table}

\subsection{Effect of Commonsense Knowledge}
\label{sec:effect_csk}
To compare the effect of commonsense knowledge, we propose another model without the CSK components. The CSK nodes and edges are discarded, and the resulting model contains only sentence nodes, global node, sentence edges, and global edges. We call this model \emph{STACK w/o CSK Nodes, Edges}. Note that this model surpasses previously reported state-of-the-art results in most datasets. To have a better understanding of how CSK helps, we compare this model with STaCK across several metrics in \cref{tab:csk_compare}. We use the following metrics for this evaluation:

\textbf{First, Last, Absolute Accuracy:} The accuracy of correctly predicting the first sentence, the last sentence, and the absolute position of any sentence in the document. \textbf{Longest Common Subsequence (LCS)} is the ratio of longest common subsequence between the predicted order and the actual order~\cite{gong2016end}. Consecutiveness is not considered necessary. The ratio is measured in percentage, and higher ratios are considered better. \textbf{Displacement} is measured by calculating the \% of sentences for which the predicted location is within distance 1 of the original location. The displacement can occur in either direction (left or right). A higher \% of this metric indicates less displacement. We denote this metric as Displacement-Window=1 or D-Win=1.
We compare the models with and without CSK in \cref{tab:csk_compare} and conclude the following:
\textbf{i)} For both CSK and w/o CSK models, predicting the correct first sentence is relatively straightforward. This is followed by correctly predicting the last sentence, and then the sentences in between.
\textbf{ii)} Incorporation of CSK always helps, except for one particular case in NSF. 
\textbf{iii)} CSK is most helpful in NeuRIPS, followed by AAN. CSK is the least helpful in NSF.
\textbf{iv)} Improvement brought by CSK varies in different degrees across the evaluation metrics. In NeuRIPS and AAN, the last sentence prediction accuracy and the absolute accuracy are improved the most after integrating CSK.

\subsection{Ablation Study}
Extending the commonsense specific analysis above (\cref{sec:effect_csk}), we further perform some ablation study on the CSK specific components of our proposed model. The results are reported in \cref{tab:results}. 
For the first ablation setting, we consider the edges with past ($p_i$) and future ($f_i$) nodes to have the same relation i.e. $r_p = r_f$. The resultant performance is slightly worse in most cases apart from the NSF dataset. The most significant drop is observed in the PMR metric of AAN, where the result is almost 2\% poorer. For NSF, this ablation setting results in improved performance, suggesting that the distinction of temporal directionality (past and future) is not essential for this dataset.

The other ablation setting corresponds to the model without the CSK components, which is the same as \emph{STaCK w/o CSK Nodes, Edges} in \cref{sec:effect_csk}. For this setting, we observe a sharp drop in performance across most of the datasets. The decrease in performance is most significant in the PMR metric of NeuRIPS and AAN. Considerable reduction in performance is also observed across various metrics in SIND and ROCStory. The ablation study with respect to CSK components coupled with the more detailed analysis in \cref{tab:csk_compare} indicates that commonsense knowledge is indeed beneficial and helps in the sentence order prediction task with varying degrees across different datasets and metrics.

We experimented with different sentence encoders and found the embeddings created by DeBERTa perform the best, followed by RoBERTa and BERT. ALBERT, on the other hand, perform the worst with around 2\% drop in $\tau$ and 4\% drop in PMR. We also experimented by removing the global node from the STaCK graph, resulting performance drop around 1\%-2\% across the datasets.

\subsection{Prediction of First and Last Sentence} 
The correct prediction of the first sentence and the last sentence is often paid more importance due to their crucial positions in a paragraph ~\cite{kumar2020deep,zhu2021neural}. We compare STaCK against BT-Sort and Constraint Graphs in \cref{tab:fl} for the task of predicting the first and last sentence correctly. Results are reported for BT-Sort and Constraint Graphs wherever available. 
First of all, we observe a common trend present in all the three methods --- the accuracy of correctly predicting the first sentence is significantly better compared to correctly predicting the last sentence. This is an interesting aspect which has been observed by other previous works as well~\cite{kumar2020deep,yin2019graph,yin2020enhancing}. 
Next, we compare the results across the three methods and find that our proposed model is significantly better than BT-Sort in predicting both the first and last sentences accurately. The difference in performance ranges from 2.8\% - 7.9\% across different datasets. We also obtain improved results over Constraint Graphs in AAN and SIND, with margins between 1.7\% - 5.9\%.

\subsection{Visualization of Learned Representations}

\subsubsection{Manifold of sentence embeddings}
\begin{figure}[ht!]
    \centering
    \includegraphics[width=0.75\linewidth]{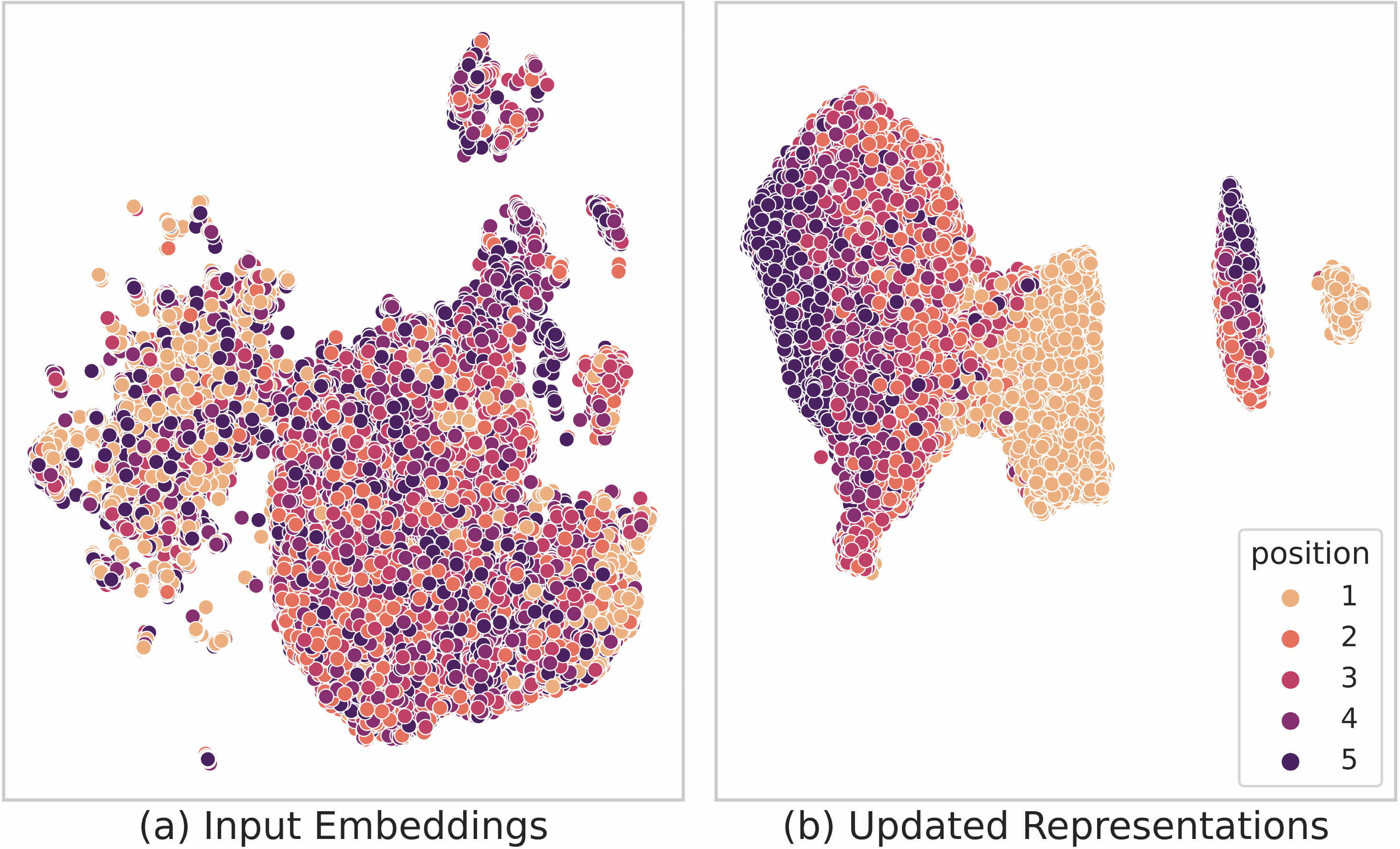}
    \caption{\footnotesize Manifold of sentence representations in ROCStory according to their position in the document. 1 and 5 indicate the first and the last sentence in a document respectively.}
    % \vspace{-0.35cm}
    \label{fig:umap}
\end{figure}
We illustrate the manifold of learned embeddings using the
% Uniform Manifold Approximation and Projection
UMAP~\cite{mcinnes2018umap} algorithm in \cref{fig:umap}. The visualization shows the test document sentences in ROCStory dataset. Sentences are colour-coded by their position (1-5) in the story. The plot on the left shows the initial sentence embeddings ($g_i$) for a non-finetuned DeBERTa model. The plot on the right shows the final node embeddings ($h_i$) in the trained graph model. Visually it is evident that the initial input embeddings do not carry much order information. However, the updated representations are much more significantly grouped together by their position. Interestingly, sentences corresponding to positions 1 (first) and 5 (last) are the most separable after the UMAP transformation. However, 
% unlike positions 1 and 5, 
sentences at positions (2-4) did not separate quite so cleanly. The results indicate that sentences appearing at the beginning and the end of a document are much easier to identify than the ones in the middle. Same conclusion can be drawn from the reported results in \cref{tab:csk_compare}.
% where it can be seen that the \emph{First} and \emph{Last} accuracy are significantly higher than the overall absolute (\emph{Abs}) accuracy. 
\begin{table}[t]
\small
\centering
\resizebox{0.65\linewidth}{!}{
\begin{tabular}{lcccc}
\toprule
Model & PMR & Abs & $\tau$ & LCS \\
\midrule
& \multicolumn{4}{c} {NeuRIPS} \\
\cline {2 - 5} B-TSort & 0.0 & 39.43 & 0.74 & 71.68 \\
STaCK & \textbf{7.69} & \textbf{44.02} & 0.74 & \textbf{74.84} \\
\hline & \multicolumn{4}{c} {AAN} \\
\cline {2 - 5} B-TSort & 0.0 & 36.86 & 0.69 & \textbf{72.01} \\
STaCK & \textbf{15.38} & \textbf{44.02} & \textbf{0.73} & 71.69 \\
\hline & \multicolumn{4}{c} {NSF} \\
\cline {2 - 5} B-TSort & \textbf{0.67} & \textbf{28.57} & \textbf{0.64} & \textbf{64.86}\\
STaCK & 0.12 & 24.42 & 0.59 & 59.67 \\
\bottomrule
\end{tabular} 
}
\caption{\footnotesize Order prediction results on NeuRIPS, ANN, and NSF datasets for documents longer than 10 sentences.}
\label{tab:long}
\end{table}
\subsubsection{Manifold of temporal knowledge} 
\begin{figure}[t]
    \centering
    \includegraphics[width=0.75\linewidth]{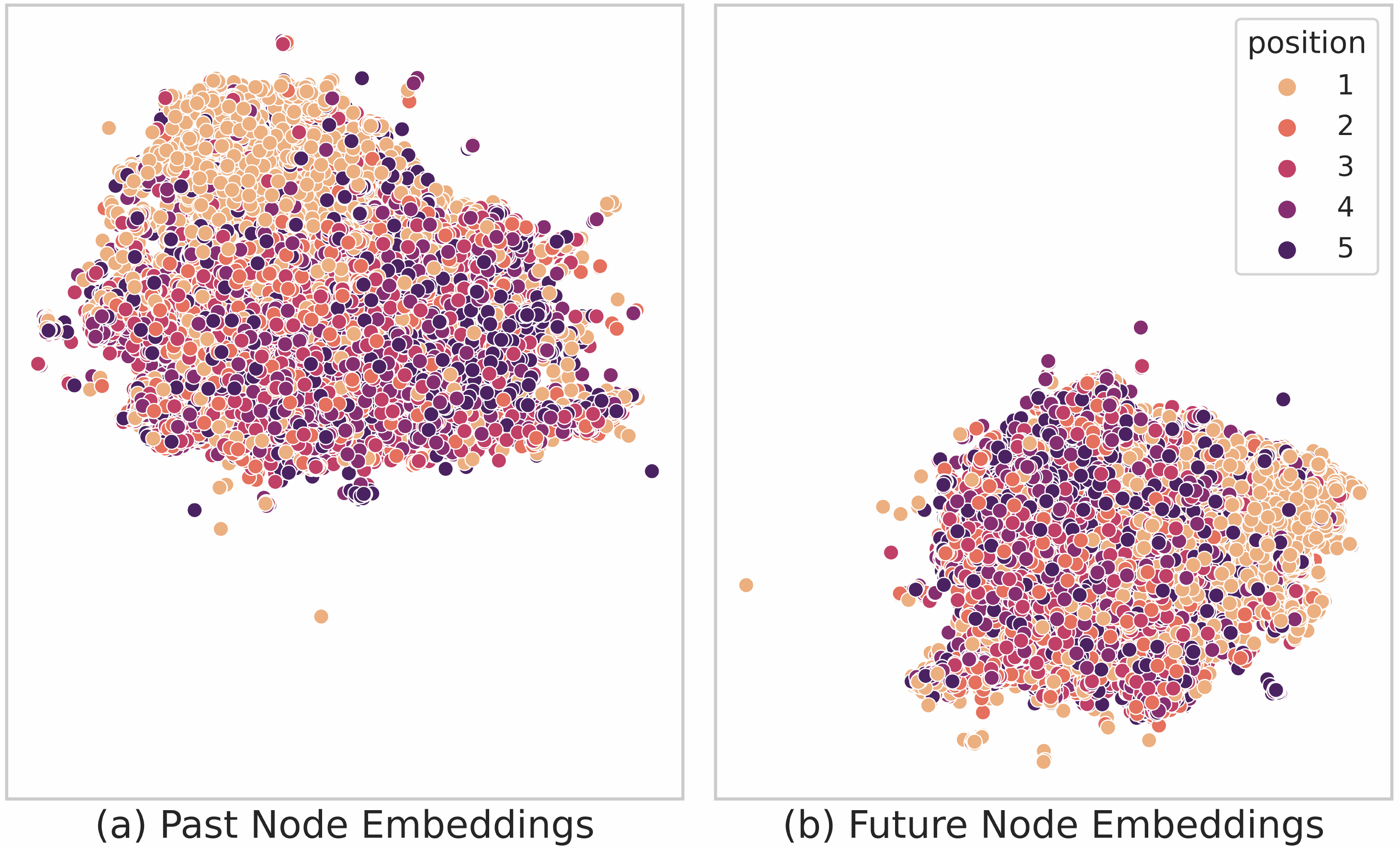}
    \caption{\footnotesize Manifold of past and future csk embeddings.}
    % \vspace{-0.35cm}
    \label{fig:umap_csk}
\end{figure}

We visualize the manifold of temporal commonsense embeddings in \cref{fig:umap_csk}. It shows the UMAP transformation of `past' and `future' node embeddings for the test sentences in ROCStory. 
% Similar to \cref{fig:umap}, the plot is colour-coded against the corresponding position of the sentence in the document.
Interestingly, embeddings corresponding to commonsense knowledge of the first sentences are grouped together more cleanly as compared to the other sentences.
% . This aspect can be noticed 
% for both `past' and `future'.
% embeddings. 
This pattern further substantiates the hypothesis, drawn from \cref{tab:csk_compare}, that the first sentences are the easiest to identify. In contrast, the embeddings corresponding to the other sentences are noisier and cannot be distinguished clearly. 

\begin{table}[t]
\centering
\small
\resizebox{\linewidth}{!}{%
\begin{tabular}{lccc}
\toprule
Gold & STaCK & \begin{tabular}[c]{@{}l@{}}STaCK \\ w/o CSK\end{tabular}  & CG \\
\midrule
 \begin{tabular}[c]{@{}l@{}}1. Bobby redid his kitchen. \\ 2. He bought a really fancy new oven. \\ 3. He couldn't wait to cook in it! \\ 4. The first time, he turned it on and smoke billowed out. \\ 5. There was something wrong with the oven!\end{tabular}                                                                       & \begin{tabular}[c]{@{}r@{}}1\\ 2\\ 3\\ 4\\ 5\end{tabular} & \begin{tabular}[c]{@{}l@{}}1\\ 3\\ 2\\ 4\\ 5\end{tabular} & 
 \begin{tabular}[c]{@{}l@{}}1\\ 2\\ 4\\ 5\\ 3\end{tabular}     \\
 \midrule
\begin{tabular}[c]{@{}l@{}}1. The family takes a trip to the local carnival. \\ 2. There are lots of rides to enjoy this year. \\ 3. There are even rides for folks as young as this small boy. \\ 4. There are also lots of games and prizes to win. \\ 5. Although some of the games seem fixed and a waste of money.\end{tabular} & \begin{tabular}[c]{@{}r@{}}1\\ 2\\ 3\\ 4\\ 5\end{tabular} & \begin{tabular}[c]{@{}l@{}}1\\ 2\\ 3\\ 4\\ 5\end{tabular} &
\begin{tabular}[c]{@{}l@{}}1\\ 2\\ 3\\ 4\\ 5\end{tabular} \\
 \midrule
 \begin{tabular}[c]{@{}l@{}}1. I heard a thud and tires screeching. \\ 2. A car went speeding by me. \\ 3. I saw a man lying on the road. \\ 4. I called 911 to report the accident. \\ 5. The police soon arrived.\end{tabular} & \begin{tabular}[c]{@{}r@{}}2\\ 1\\ 3\\ 4\\ 5\end{tabular} & \begin{tabular}[c]{@{}l@{}}2\\ 3\\ 4\\ 1\\ 5\end{tabular} &   
 \begin{tabular}[c]{@{}l@{}}1\\ 3\\ 2\\ 5\\ 4\end{tabular} \\
\bottomrule
\end{tabular}%
}
\caption{\footnotesize Case studies in ROCStory and SIND. STaCK produces more accurate predictions by using commonsense and contextual knowledge from the documents.}
\label{tab:case_study}
\end{table}
\subsection{Order Prediction in Longer Documents}
\noindent We report order prediction results for documents having more than ten sentences in \cref{tab:long}. The Constraint Graphs paper does not report this result, and thus we compare STaCK with the BT-Sort method. We report results only in NeuRIPS, AAN, and NSF, as SIND and ROCStory have exactly five sentences in all documents.
From the results, we conclude that STaCK is significantly better than BT-Sort for long documents in NeuRIPS and AAN. The \emph{perfect match ratio}, and \emph{absolute accuracy} are several percentage point higher for STaCK compared to BT-Sort. For NSF, both the models perform very poorly in the PMR metric, with scores lesser than 1\%. However, BT-Sort has superior performance compared to STaCK across the other metrics. Note that the overall result of BT-Sort was worse compared to STaCK (\cref{tab:results}). Results from \cref{tab:long} and \cref{tab:results} suggest that, BT-Sort is better for longer documents and STaCK is better for shorter documents in NSF.

\subsection{Case Studies}
We report a few case studies in \cref{tab:case_study}. Gold order of three documents from ROCStory and SIND dataset are shown on the left. The columns on the right depicts the order predicted by our framework with and without CSK, and the Constraint Graphs (CG) model. STaCK predicts the sentence order most accurately, whereas STaCK w/o CSK often swaps absolute positions or shifts consecutive sentences. CG predicts the first sentence correctly in all cases, but suffers from predicting contextual discrepancies. For instance, \emph{He couldn't wait to cook in it!} is predicted after \emph{The first time, he turned it on and smoke billowed out}. In the third example, temporal commonsense around the event \emph{I called 911 to report the accident} is aligned to the event \emph{The police soon arrived} through the relation \emph{isBefore} in COMET. Such commonsense knowledge helps in predicting the entire order correctly. We note that CG predictions for this example are displaced within window 1, with \emph{I called 911 to report the accident} and \emph{The police soon arrived} having wrong relative order. Such instances of the importance of document information and CSK are prevalent throughout the dataset.

\subsection{Effect of COMET}
We experimented by adding other temporal and causal commonsense relations in COMET such as \emph{causes}, \emph{as-a-result}, \emph{desires}, \emph{requires} as nodes to the proposed graph in STaCK. However, they did not result in any significant performance improvements. We posit this could be due to the fact that there exists a large overlap between the generated output of \emph{isBefore}, \emph{isAfter} and the four relations as mentioned above. Nonetheless, we think all types of CSK relations available in COMET can be used in the task. The graph structure to accommodate those additional CSK is left as future work.

\subsection{Choice of Transformer Encoder}
\label{sec:encoder_choice}
The choice of the transformer encoder plays a crucial role in the sentence order prediction task. Several choices of transformer-based models are available, such as BERT~\cite{devlin2019bert}, RoBERTa~\cite{liu2019roberta}, ALBERT~\cite{lan2019albert}, DeBERTa~\cite{he2020deberta}, etc. Different objective functions are used to pre-train these models, which directly affects how these models perform on the downstream sentence ordering task. In particular, BERT is pre-trained with the masked language modelling (MLM) and the next sentence prediction (NSP) objective. RoBERTa and DeBERTa models are pre-trained only with the MLM objective. ALBERT model is pre-trained with the masked language modelling (MLM) and a sentence order prediction (SOP) objective. 

BT-Sort and Constraint Graphs are the current state-of-the-art models which use a BERT encoder. Both models are of pair-wise nature (\cref{sec:related_work}), that first predict the relative order between each pair of sentences in the document. The final order is then inferred from all the relative orders. The relative order prediction is performed by concatenating the sentence pairs with a \emph{<SEP>} token and then passing through BERT encoder. This setting directly aligns with the NSP objective of BERT, and is capable of achieving state-of-the-art results. However, as reported in ~\cite{zhu2021neural}, replacing the BERT encoder with a RoBERTa encoder results in poorer performance because of the absence of the NSP objective. Interestingly, using ALBERT encoder also results in a performance drop, even though ALBERT was pre-trained with a sentence order prediction objective (albeit rather differently). Furthermore, we also experimented by replacing the the BERT encoder of BT-Sort with DeBERTa and found that the performance does not surpass the reported results of BERT. Our proposed model is also a pair-wise model. However, the graph-based encoding technique is different from the commonly used sentence pair concatenating method. We found that for our graph model, sentence embeddings created by DeBERTa perform the best, followed by RoBERTa and BERT. Sentence embeddings produced by ALBERT perform the worst.

\section{Conclusion}
%\vspace{-0.4cm}
In this work, we presented STaCK, a framework that uses Relational Graph Convolutional Network (RGCN) to model document-level contextual information and temporal commonsense knowledge for sentence order prediction. In the graph network, the edge classification objective was applied for pair-wise relative order prediction of the sentence pairs. This was followed by a topological sorting for the final order prediction of the sentences. STaCK achieves state-of-the-art results in several benchmark datasets.

\section*{Acknowledgments}
This work is supported by the AcRF MoE Tier-2 grant titled: ``CSK-NLP: Leveraging Commonsense Knowledge for NLP'' and the A*STAR under its
RIE 2020 Advanced Manufacturing and Engineering (AME) programmatic grant, Award No. -
A19E2b0098.
\bibliography{anthology,custom}
\bibliographystyle{acl_natbib}

% \appendix

% \section{Example Appendix}
% \label{sec:appendix}

% This is an appendix.

\end{document}